\documentclass[letterpaper,journal,twoside]{IEEEtran}
\usepackage{times}  
\usepackage{helvet}  
\usepackage{courier}  
\usepackage[hyphens]{url}  
\usepackage{graphicx} 
\usepackage{caption} 
\usepackage{algorithm}
\usepackage{algorithmic}
\usepackage{newfloat}
\usepackage{subcaption}
\usepackage{booktabs}
\usepackage{amsmath}
\usepackage{multirow}
\usepackage{color}
\usepackage[compress]{cite}
\usepackage{colortbl}  
\usepackage{xcolor}
\begin{document}
\title{\textbf{Learning Rollout from Sampling:\\ An R1-Style Tokenized Traffic Simulation Model}}

\author{
$\text{Ziyan~Wang}^{1}$,
$\text{Peng~Chen}^{1}$,
$\text{Ding~Li}^{1\ddag}$,
$\text{Chiwei~Li}^{1}$,
$\text{Qichao~Zhang}^{2}$,
$\text{Zhongpu~Xia}^{2}$
and $\text{Guizhen~Yu}^{1}$

\thanks{
Manuscript received: November 5, 2025; Revised: January 24, 2026; Accepted: March 5, 2026.}
\thanks{
This paper was recommended for publication by Editor Tamim Asfour upon evaluation of the Associate Editor and Reviewers’ comments. This work was supported by National Natural Science Foundation of China (NSFC) under Grant 62503033 and 52272327.}
\thanks{$^{1}$ Ziyan Wang, Peng Chen, Ding Li, Chiwei Li, Guizhen Yu  are with State Key Laboratory of Intelligent Transportation System, Key Laboratory of Autonomous Transportation Technology for Special
Vehicles, Ministry of Industry and Information Technology, School of Transportation Science and Engineering, Beihang University,
Beijing 100191, China. {\tt\footnotesize liding@buaa.edu.cn}}
\thanks{$^{2}$ Qichao Zhang, Zhongpu Xia are with State Key Laboratory of Multimodal Artificial Intelligence Systems, Institute of Automation, Chinese Academy of Sciences, Beijing,
100190, China. {\tt\footnotesize zhangqichao2014@ia.ac.cn}}
\thanks{$^{\ddag}$ Corresponding author.}
\thanks{Digital Object Identifier (DOI): see top of this page.}
}
\markboth{IEEE ROBOTICS AND AUTOMATION LETTERS. PREPRINT VERSION. ACCEPTED MARCH 2026}%
{Wang \MakeLowercase{\textit{et al.}}: R1-Style Tokenized Traffic Simulation}
\maketitle
\begin{abstract}
Learning diverse and high-fidelity traffic simulations from human driving demonstrations is crucial for autonomous driving evaluation. The recent next-token prediction (NTP) paradigm, widely adopted in large language models (LLMs), has been applied to traffic simulation and achieves iterative improvements via supervised fine-tuning (SFT). However, such methods limit active exploration of potentially valuable motion tokens, particularly in suboptimal regions.
Entropy patterns provide a promising perspective for enabling exploration driven by motion token uncertainty. Motivated by this insight, we propose a novel tokenized traffic simulation policy, R1Sim, which represents an initial attempt to explore reinforcement learning based on motion token entropy patterns, and systematically analyzes the impact of different motion tokens on simulation outcomes. Specifically, we introduce an entropy-guided adaptive sampling mechanism that focuses on previously overlooked motion tokens with high uncertainty yet high potential. We further optimize motion behaviors using Group Relative Policy Optimization (GRPO), guided by a safety-aware reward design.
Overall, these components enable a balanced exploration–exploitation trade-off through diverse high-uncertainty sampling and group-wise comparative estimation, resulting in realistic, safe, and diverse multi-agent behaviors. Extensive experiments on the Waymo Sim Agent benchmark demonstrate that R1Sim achieves competitive performance compared to state-of-the-art methods.
\end{abstract}


\section{Introduction}

\IEEEPARstart{S}{mart} traffic simulation is essential for autonomous driving validation \cite{yang2025drivearena, feng2021intelligent,luo2022gamma} and the continuous improvement of autonomous driving policies within a safe, scalable environment. Drawing inspiration from the success of large language models (LLMs) in natural language processing \cite{achiam2023gpt}, the smart traffic simulation paradigm can be formulated as a multi-agent policy learning task with autoregressive modeling, with agent trajectories tokenized into discrete motion representations and trained through imitation learning-based next token prediction (NTP) for motion generation. State-of-the-art (SOTA) frameworks like SMART \cite{wu2024smart} and CATK \cite{zhang2025closed} have demonstrated exceptional scalability in real-time multi-agent motion generation, establishing pretrained foundation models for complex traffic simulation scenarios.

\begin{figure}[!t]
	\centering
		\scriptsize
		\includegraphics*[width=3.5in]{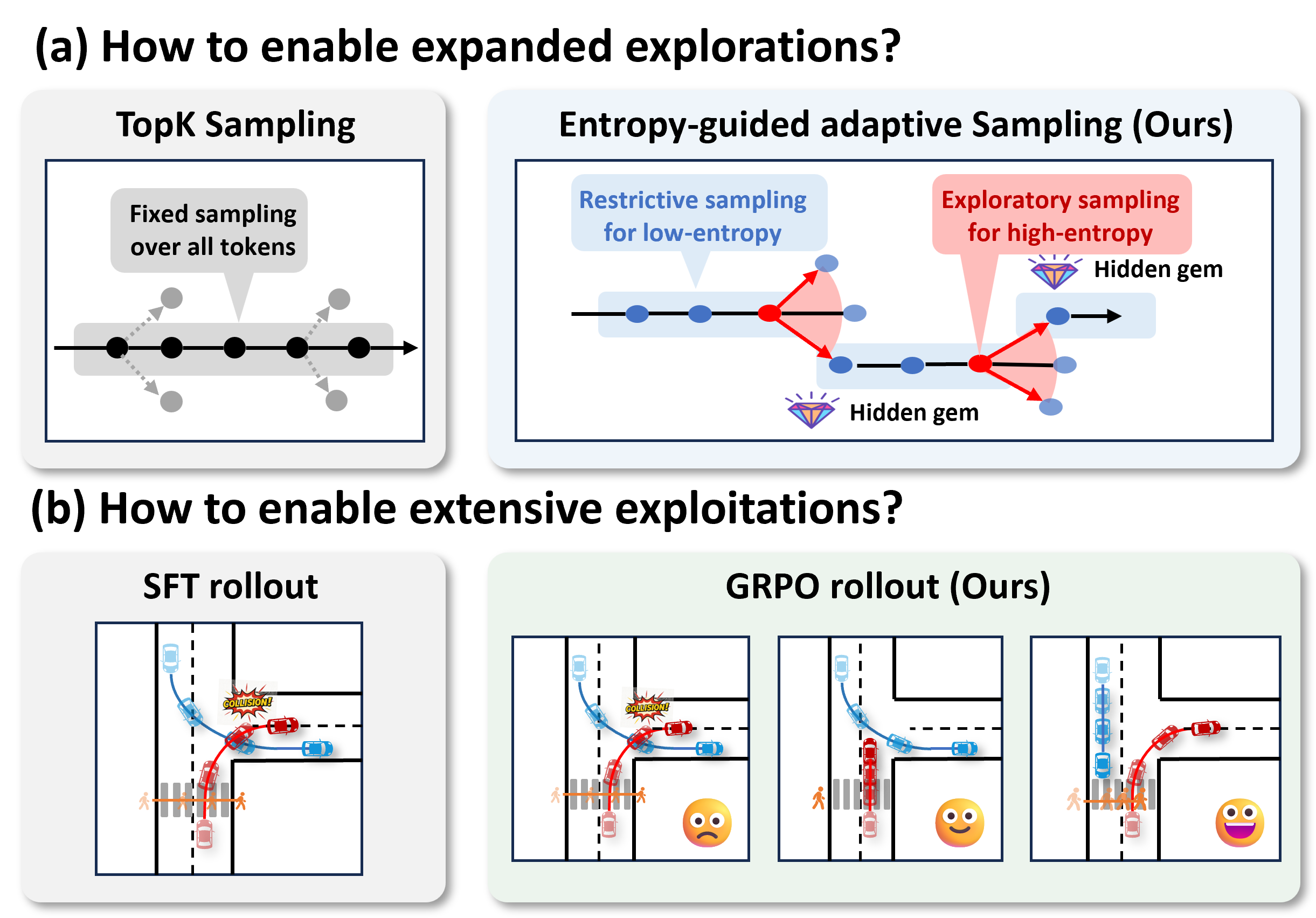}
    \caption{\textbf{{The motivation of R1Sim.}} (a) Exploration: Compared with Top-K sampling \cite{wu2024smart}, our entropy-guided adaptive sampling explores more high-entropy motion tokens.
(b) Exploitation: Compared with SFT \cite{zhang2025closed}, our refined GRPO estimates group-wise advantages and selects the optimal scenario.}
		\label{fig:motivation}
\vspace{-1em}
\end{figure}

Despite promising motion generation stability, existing tokenized motion models still fail to handle these two key challenges: {\textit{1) how to enable adaptive exploration to sample multiple plausible motion scenarios.} SOTA approaches \cite{wu2024smart,zhang2025closed} implement the Top-K sampling strategy \cite{fan2018hierarchical} to select a fixed number of motion tokens for simulation rollout, as shown in Fig.~\ref{fig:motivation}~(a). The rigid strategy over-prioritizes high-probability motion tokens from the vocabulary while neglecting potentially valuable “hidden gem" behaviors in the token vocabulary, particularly detrimental in interactive scenarios, where diverse motion outcomes are essential. Once the exploration space is established, \textit{2) how to enable effective exploitation to optimize realism and safety of multi-agent motion behaviors.} Existing optimization methods \cite{lin2025revisit,zhang2025closed} such as supervised fine-tuning (SFT), often employ winner-takes-all approaches that force generated states to match expert demonstrations. However, as shown in Fig. ~\ref{fig:motivation}~(b), over-reliance on potentially suboptimal ground truth may perpetuate unsafe behaviors.} Thus, it is crucial to balance well between exploration and exploitation to discover “hidden gem” behaviors that best align with human-preferred motion tokens.

To address these challenges, we introduce R1Sim, a novel framework that pioneers the integration of motion token entropy dynamics into reinforcement learning for autonomous traffic simulation. We observe that entropy variations in tokenized motion simulation models effectively capture distributional modeling uncertainty \cite{cui2025entropymechanismreinforcementlearning,gray2011entropy,meister2025locallytypicalsampling}: low entropy typically corresponds to predictable, inertial maneuvers, while high entropy signals complex, multi-modal driving intentions. Motivated by this, R1Sim establishes a balanced exploration–exploitation mechanism tailored for traffic generation. For exploration, we propose an entropy-guided adaptive sampling strategy. Unlike conventional fixed Top-K sampling, our proposed sampling mechanism dynamically relaxes constraints under high uncertainty to uncover physically plausible yet low-probability “hidden gem” behaviors that are often missed by standard greedy decoding. For exploitation, to strictly align these diverse proposals with human safety and realism preferences, we introduce a refined Group Relative Policy Optimization (GRPO)\cite{guo2025deepseekr1,guo2024deepseekmath}. By contrasting token-level advantages within rollout groups and incorporating a fine-grained, safety-aware reward function, our method systematically reinforces optimal decision-making. This approach overcomes the limitations of both traditional RL value estimation and SFT, enabling progressive discrimination between optimal and suboptimal motion patterns as illustrated in Fig. \ref{fig:motivation}. 
Our contributions are listed as follows:

\begin{itemize}
    \item We propose a human-preferred motion simulation framework, R1Sim, by balancing exploration-exploitation within a next token prediction pretraining paradigm.
    \item We introduce an entropy-guided adaptive sampling strategy to enhance exploration by dynamically selecting high-uncertainty yet potentially optimal motion tokens.
    \item We develop a GRPO-refined method with safety-aware rewards to effectively exploit high-quality behaviors.

\end{itemize}
\section{Related Work}

\subsection{Tokenized Traffic Simulation}
Autoregressive modeling has demonstrated strong expressive power and flexibility in modeling complex sequential data.
Recent advances in this paradigm have substantially improved the realism of traffic simulation \cite{wu2024smart, philion2024trajeglishtrafficmodelingnexttoken,seff2023motionlm}. These methods discretize complex traffic behaviors into sequential motion tokens to enable long-horizon motion generation \cite{zhou2024behaviorgpt}. Despite demonstrating promising simulation results, these models suffer from covariate shift due to distributional discrepancies between training and inference states step by step \cite{de2019causal}.

To mitigate this limitation, recent works have investigated supervised fine-tuning. For instance, CATK \cite{zhang2025closed} establishes a closed-loop fine-tuning paradigm, using SMART as its pretrained foundation to generate high-fidelity tokenized rollouts that closely match ground truth trajectories. From another aspect, UniMM \cite{lin2025revisit} augments closed-loop samples from the policy itself to mitigate covariate shift. 

However, these methods face two inherent limitations in NTP paradigms: \textbf{1)} Rigid fixed sampling strategies create a scalability bottleneck for diverse motion generation. \textbf{2)} Reliance on ground truth constrains exploration in sub-optimal regions, hindering the discovery of tokens that better align with the underlying driving logic.

\subsection{Reinforcement Learning Fine-tuning}

Great breakthroughs of LLMs in aligning with human preference via reinforcement learning fine-tuning (RLFT) have inspired a parallel evolution in autonomous driving. Initial works such as BC-SAC \cite{lu2023imitation} utilize simple rewards to enhance imitation learning, while subsequent research \cite{huang2024gen,li2025finetuning} adopts Reinforcement Learning from Human Feedback (RLHF) to unlock diverse driving behaviors. Notable examples include TrajHF \cite{li2025finetuning}, which refines multi-modal motion generation results by incorporating multi-conditional denoiser, and Carplanner \cite{zhang2025carplanner}, which optimizes planning via Proximal Policy Optimization (PPO) \cite{schulman2017proximal}. Nevertheless, the computational overhead of PPO's value function estimation poses challenges for real-time control. To address this, DeepSeek-R1 \cite{guo2025deepseekr1} introduces Group Relative Policy Optimization (GRPO), which optimizes policies through group-wise outcomes without requiring a critic network. This paradigm has validated its efficacy in handling complex generative tasks, ranging from task decomposition \cite{parmar2025plan}, language reasoning \cite{dou2025plan} and text-to-image generation. Although originally developed for large language models, GRPO is well suited for autonomous driving, as both domains adopt an autoregressive formulation and benefit from group-wise comparison over multiple sampled candidates. Recent models such as AlphaDrive \cite{jiang2025alphadrive} and Plan-R1\cite{tang2025plan} stabilize long-horizon motion planning.  In this paper, we pioneer the investigation of motion token entropy dynamics into GRPO for traffic simulation.

\section{Preliminary}
\subsection{Tokenized Traffic Simulation Formulation}
Our objective is to develop a traffic simulation policy $\pi_{\theta}$ to produce realistic and safe behaviors of all agents involved in driving scenarios, where $\theta$ denotes the learnable parameters. Given the map information $\mathcal{M}$, the policy models the probability distribution of the motion states $\{S_{0},...,S_{T}\}$ across the simulation horizon $T$ in an autoregressive manner. The probability distribution can be described by 
\begin{equation} \label{eq:rollout_generation_abstraction}
p(S_{0}, \dots, S_{T} \mid \mathcal{M}, \pi_\theta) \propto \prod_{t=0}^T p(S_{t} \mid S_{<t}, \mathcal{M}; \pi_\theta),
\end{equation}
where $p(S_{t}\mid S_{<t}, \mathcal{M}, \pi_\theta)$
denotes the probability distribution of motion state $S_{t}$ conditioned on the prior states $S_{<t}$ and the map information $\mathcal{M}$. At each time step $t$, the motion state $S_{t}$ represents the joint states of all agents, i.e., $S_t = \{s_{t,j} | j \in [1, N_{agent}]\}$, and $N_{agent}$ denotes the number of all agents in the traffic scenarios.

Built upon the NTP paradigm, the traffic simulation policy models the probability distribution over the discrete token vocabulary $V$, with each discrete token $c \in V$ denoting a short-term maneuver command. At each timestep $t$, the joint action $C_t$ of all agents can be sampled from  
\begin{equation}
    \pi_\theta(C_t \mid S_{<t},\mathcal{M}) = \prod_{j=1}^{N_\text{agent}}  \pi_\theta(c_{t,j} \mid S_{<t},\mathcal{M}).
\end{equation}

\subsection{Motion Token Entropy Definition}
In this work, we define \emph{motion token entropy} as the information entropy of the vocabulary distribution produced by the decoder of the tokenized traffic simulation model under a given input condition $(S_{<t}, \mathcal{M})$. At time step $t$, the model outputs a categorical distribution $p_t(\cdot)$ over the motion token vocabulary $V$, reflecting the uncertainty in selecting the next motion token, as follows:

\begin{equation}
{p_{t}(c)=\pi_\theta\left(c \mid S_{<t}, \mathcal{M}\right), c \in V.}
\end{equation}
{The motion token entropy at time step $t$ is then computed as the Shannon entropy of this distribution:}
\begin{equation}
\label{eq:entropy}
{H_{t}=-\sum_{c \in V} p_{t}(c)\log p_{t}(c).}
\end{equation}

\begin{figure}[!t]
	\centering
		\scriptsize
		\includegraphics*[width=3.5in]{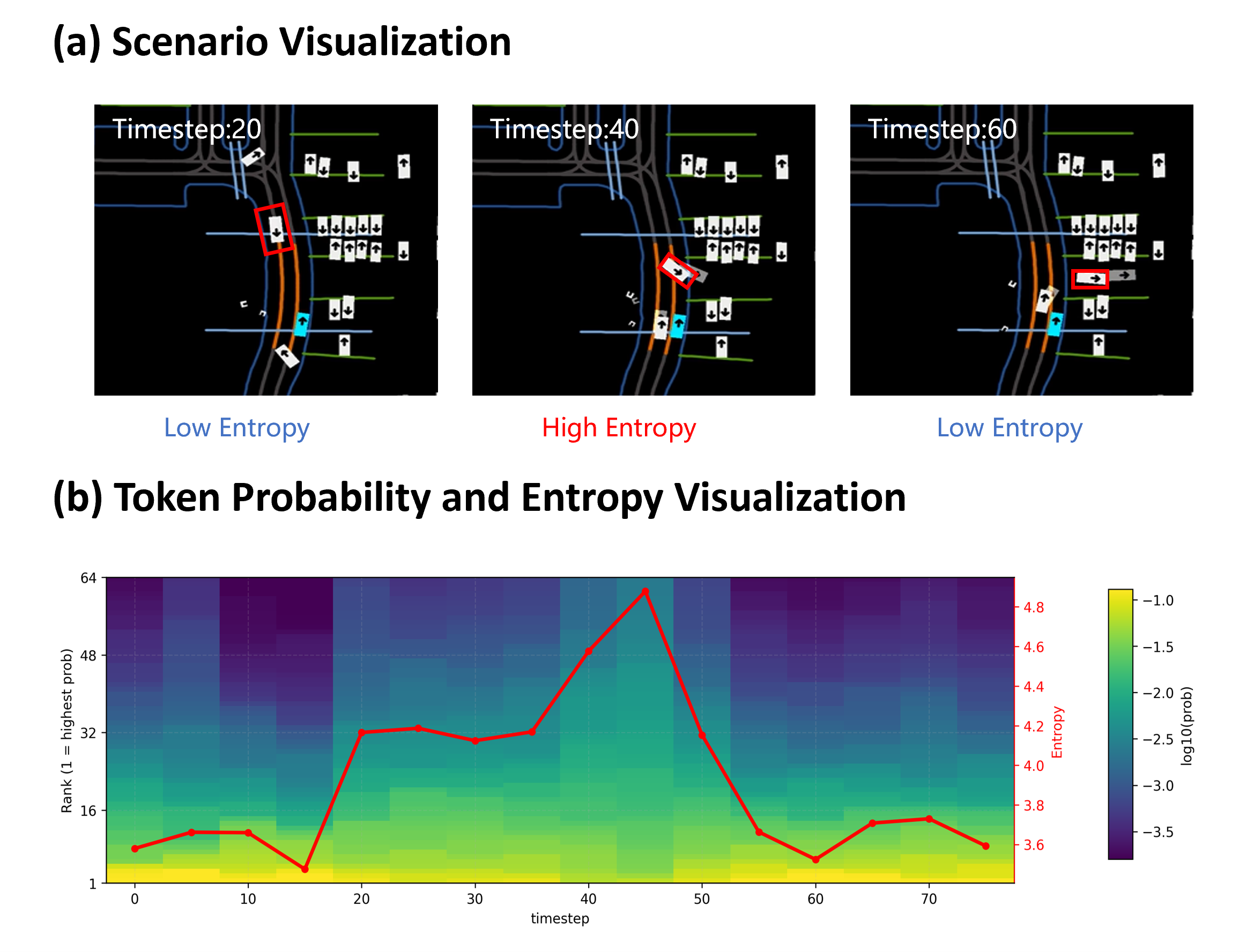}
		\caption{\textbf{Temporal evolution of token entropy and its role in characterizing scene uncertainty.} (a) visualizes representative low- and high-entropy motion patterns at different time steps, highlighting the interested vehicle in red. 
(b) shows the temporal evolution of the ranked token probability distribution and the corresponding motion token entropy, where higher entropy coincides with a flatter distribution.}
		\label{fig:entropy}
\vspace{-1em}
\end{figure}

Fig.~\ref{fig:entropy} shows the role of motion token entropy in shaping temporal evolution of multi-agent traffic scenes under an autoregressive simulation framework. Lower motion token entropy indicates that the model assigns high confidence to a small set of motion tokens, corresponding to deterministic motion patterns. Meanwhile, higher motion token entropy emerges in complex and interactive phases (e.g., around T=40), where multiple plausible motion choices exist, reflecting increased uncertainty. 
This motion token entropy signal provides a principled, time-varying measure of scene uncertainty and thereby enhances comprehensive explorations of potentially optimal behaviors.

\section{Method}

\begin{figure*}[htbp]
	\centering
	\begin{center}
		\scriptsize
		\includegraphics*[width=6.9in]{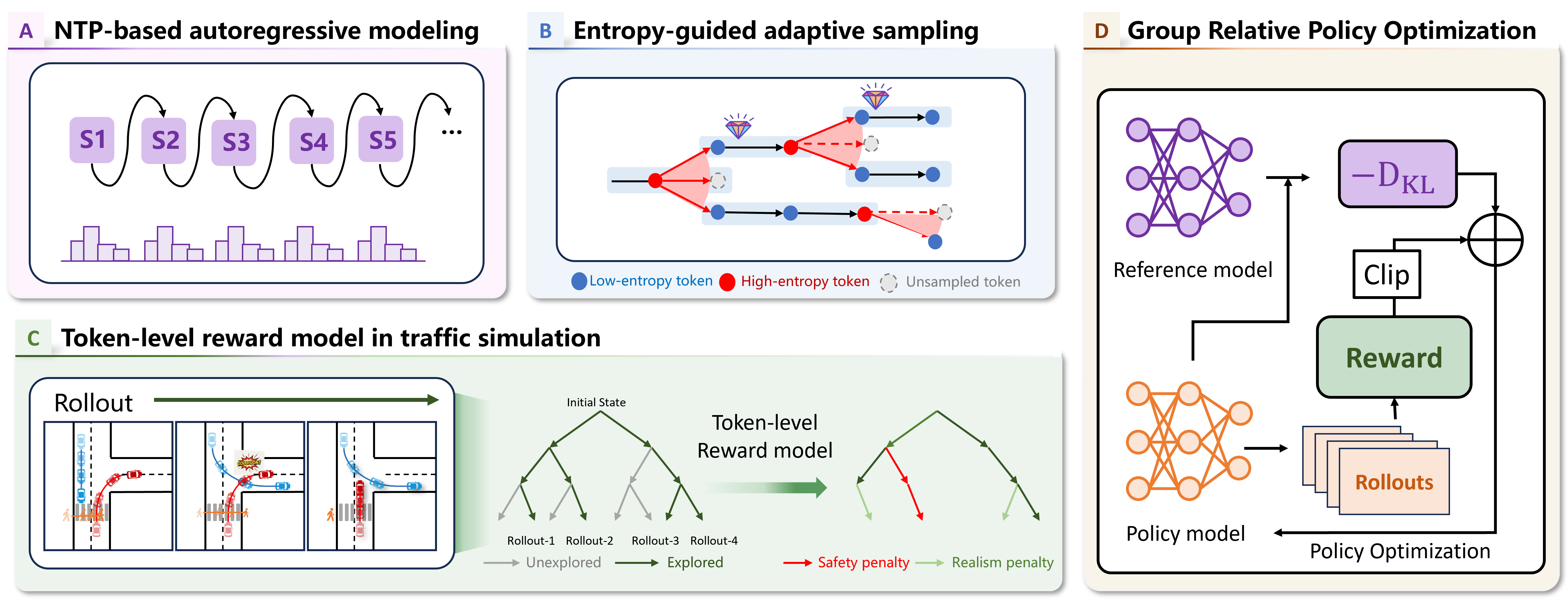}
		\caption{\textbf{{An overview of R1Sim framework.}} {Our framework follows (A) an NTP-based autoregressive formulation for sequential motion token generation. Exploration is facilitated by (B) an entropy-guided adaptive sampling strategy that allocates sampling budget according to token entropy, while exploitation is guided by (C) a token-level reward model operating in traffic simulation. Policy network is optimized using (D) GRPO, enabling learning through group relative advantage estimation.}}
		\label{fig:framework}
	\end{center}
\vspace{-1em}
\end{figure*}

As illustrated in Fig. \ref{fig:framework}, R1Sim follows an NTP-based autoregressive framework for sequential motion token generation. Given the current scene context, the policy first estimates token-level uncertainty via entropy and performs entropy-guided adaptive sampling to generate diverse candidate rollouts. These rollouts are then evaluated using a token-level, safety-aware reward defined in the traffic simulation environment. Finally, the policy is optimized with GRPO, leveraging group-wise relative advantages and KL regularization to reinforce human-preferred motion behaviors.
\subsection{Entropy-Guided Adaptive Sampling}
Instead of using a fixed Top-K sampling strategy during the rollout \cite{fan2018hierarchical}, we are inspired by the novel perspective of token entropy patterns in LLMs \cite{wang20258020rulehighentropyminority} and newly propose an entropy-guided adaptive sampling strategy to tailor the token sampling from model uncertainties. The model uncertainties can be measured by the entropy of policy distribution, with the lower entropy measuring a peak distribution from which the policy commonly samples, as well as the higher entropy measuring a flatter distribution where the diverse sampling would be focused more. 

Building on this, we dynamically adjust the sampling range parameters $K_t$ based on entropy, thereby achieving an optimal balance between exploration and exploitation during policy sampling.  
\begin{equation}
C_{t} \sim \operatorname{Top}^{K_t}\left(\pi_\theta\left(c \mid S_{<t}, \mathcal{M}\right)\right),
\end{equation}
where $\operatorname{Top}^{K_t}(\cdot)$ denotes the sampling operator with the adaptive sampling parameter $K_t$. The adaptive sampling parameter $K_t$ is adopted to sample diverse motion tokens $C_t$ of all agents from the categorical distribution $\pi_\theta\left( \cdot \mid S_{<t}, \mathcal{M}\right)$ over the token vocabulary $V$.

Specifically, we compute the adaptive sampling parameter $K_{t}$ guided by the entropy of policy distribution:
\begin{equation}
K_{t}=k_{min}+\frac{k_{max}-k_{min}}{1+e^{-H_t}},
\label{eq:k}
\end{equation}
where $k_{min}$ and $k_{max}$ represent the minimum and maximum threshold values of the sampling number. The adaptive sampling parameter $K_t$ can be formulated as a monotonically increasing function of the entropy $H_{t}$ as shown in Eq. (\ref{eq:k}). This strategy leverages model uncertainties, measured by the entropy $H_t$, to well balance exploration-exploitation. Especially, we perform exploitations on the highest-probability motion tokens to ensure accurate decision-making when encountered with the lower entropy. Once the higher entropy arises, the proposed sampling strategy promotes broader explorations within a wide range of motion tokens.

\subsection{Group Relative Policy Optimization}
As shown in Fig. \ref{fig:framework}, to align generated trajectories with safety and realism constraints, we propose R1Sim, which fine-tunes the motion generator using a refined GRPO. We decompose the evaluation into fine-grained token-level rewards and optimize the policy via a specialized advantage formulation that enhances stability in traffic simulation. This optimization process encourages the model to discover and reinforce beneficial motion tokens originating from the high-entropy uncertainty states. We present the pseudo-code of GRPO in Algorithm 1.
\begin{algorithm}[t]
  \hrule
  \caption*{\textbf{Algorithm 1}: GRPO for tokenized traffic model}
  \begin{algorithmic}[1]
    \REQUIRE Pretrained policy $\pi_\theta$, token vocabulary $V$
    \ENSURE Finetuned policy model $\pi_\theta$
    \STATE Init reference model $\pi_{ref} \leftarrow \pi_\theta$
    \REPEAT 
    \STATE Sample a traffic scenario $\{ \hat{S}_{0}, \mathcal{M} \}$
      \STATE Init rollout state $S_0 = \hat{S}_0$
      \FOR{$i = 1$ \TO $N_{\text{rollout}}$}
        \FOR{$t = 0$ \TO $N_{\text{step}}$}
          \STATE Sample action $C_{i,t} \sim \pi_{\theta}(\cdot | S_{i,<t}, \mathcal{M})$ from $V$
        \ENDFOR
      \ENDFOR
      \STATE Compute rewards $\{r_i\}_{i=1}^{N_{\text{rollout}}}$ (Eq.~\ref{eq:reward_function})
      \STATE Compute group relative advantages $a_{i,t}$ (Eq.~\ref{eq:advantage_function})
      \STATE Update $\pi_\theta$ by minimizing the GRPO loss (Eq.~\ref{eq:grpo_loss})
      \STATE Update old policy $\pi_{\theta_{\text{old}}} \leftarrow \pi_{\theta}$
    \UNTIL{convergence or max iterations}
  \end{algorithmic}
  \label{alg:grpo_Fine-tuning}
\end{algorithm}

We define the safety-aware reward $r_{i,t}$ of a rollout $o_i$, balancing collision avoidance and trajectory fidelity:
\begin{equation}
\label{eq:reward_function}
    r_{i,t}=  r_{i,t}^{\text{safe}} \cdot r_{i,t}^{\text{dis}}.
\end{equation}
The safety reward $r_{i,t}^\text{safe}$ penalizes collisions detected via the Separating Axis Theorem (SAT):
\begin{equation}
r_{i,t}^\text{safe} =
\left\{
\begin{array}{cl}
-1 & \text{if $\operatorname{SAT}(C_{i,t})$ } \\
1  & \text{else}
\end{array}
\text{where $C_{i,t} \in o_i$}.
\right.
\end{equation}
The realism reward $r_{i,t}^{\text{dis}}$ uses a negative exponential kernel to penalize deviations from the ground truth $y_t$, prioritizing: “hidden gems" even when they diverge from the ground truth:
\begin{equation}
r_{i,t}^\text{dis} = \exp\left(-\alpha \left| S_{i,t} - y_t \right| \right).
\end{equation}

To better align the policy with human-preferred simulation principles, we employ GRPO as our reinforcement learning algorithm by computing relative advantages across diverse sampled motion token groups. Intuitively, instead of forcing the policy to match a single reference trajectory, this formulation compares multiple candidate motion tokens generated under the same traffic scene and reinforces those leading to safer and more realistic interactions relative to others. During fine-tuning, we initialize the pretrained model as the reference policy $\pi_{ref}$ and optimize the motion generator as the trainable policy $\pi_{\theta}$. At each timestep, GRPO samples diverse motion states from the old policy $\pi_{\theta_{old}}$. The policy $\pi_{\theta}$ is updated by minimizing the loss function:

\begin{equation}
\label{eq:grpo_loss}
\begin{split}
{L(\theta) =} & {-\mathbf{E}_{S_0\sim\mathcal{D},\{o_i\}_{i=1}^{N_\text{rollout}}\sim \pi_{\theta_\text{old}} \left(O\mid \{\hat{S}_{0:T}, \mathcal{M}\}\right)} }\\
& {\quad \frac{1}{N_{\text{rollout}}} \sum_{i=1}^{N_{\text{rollout}}} \sum_{t=0}^{T} \left\{ \min \left[ \rho_{i,t} {A}_{i,t}, \right. \right.} \\
& {\quad \left. \left. \operatorname{clip}\left(\rho_{i,t}, 1-\epsilon, 1+\epsilon\right) {A}_{i,t} \right] - \beta_{\text{KL}} D_{\text{KL}}(\pi_{\text{ref}} \| \pi_\theta) \right\},}
\end{split}
\end{equation}
\begin{table*}[htbp]
\caption{Results on the WOMD Test Dataset}
  \centering
  \begin{tabular}{l|cccccc}
    \toprule
    Method & 
    \begin{tabular}[c]{@{}c@{}}RMM (↑)\end{tabular} &
    \begin{tabular}[c]{@{}c@{}}Kinematic L.(↑)\end{tabular} &
    \begin{tabular}[c]{@{}c@{}}Interactive L.(↑)\end{tabular} &
    \begin{tabular}[c]{@{}c@{}}Map-based L.(↑)\end{tabular} &
    \begin{tabular}[c]{@{}c@{}}Collision L.(↑)\end{tabular} &
    \begin{tabular}[c]{@{}c@{}}Offroad L.(↑)\end{tabular} \\
    \midrule
    UniMM \cite{lin2025revisit}  & 0.7684 & \textbf{0.4913} & 0.8101 & \underline{0.8737} & 0.9679 & 0.9506 \\
    DRoPE \cite{zhao2025dropedirectionalrotaryposition} & 0.7625 & 0.4779 & 0.8065 & 0.8685 & 0.9607 & 0.9443 \\
    SMART-large \cite{wu2024smart}  & 0.7614 & 0.4786 & 0.8066 & 0.8648 & 0.9632 & 0.9403 \\
    KiGRAS \cite{zhao2024kigras} & 0.7597 & 0.4691 & 0.8064 & 0.8658 & 0.9617 & 0.9431 \\
    BehaviorGPT \cite{zhou2024behaviorgpt}  & 0.7473 & 0.4333 & 0.7997 &  0.8593 & 0.9537 & 0.9349 \\
    GUMP \cite{hu2024solving}  & 0.7431 & 0.4780 & 0.7887 & 0.8359 & 0.9403 & 0.9028 \\
    \midrule
    SMART-tiny \cite{wu2024smart} & 0.7591 & 0.4759 & 0.8039 & 0.8632 & 0.9601 & 0.9401 \\
    \textbf{SMART-tiny w/R1Sim}  & 0.7675 & 0.4894 & 0.8105 & 0.8710 & \textbf{0.9718} & 0.9490 \\
    CATK \cite{zhang2025closed}  & \underline{0.7687} & 0.4909 & \underline{0.8105} & \textbf{0.8739} & 0.9707 & \textbf{0.9517} \\
    \textbf{CATK w/R1Sim} & \textbf{0.7688} & \textbf{0.4913} & \textbf{0.8107} & 0.8735 & \underline{0.9713} & \underline{0.9516} \\
    \bottomrule
  \end{tabular}
  \vspace{2mm}
  \\
  {\footnotesize The best and second results are highlighted in \textbf{bold} and \underline{underline}. L. denotes likelihood.}
  \label{tab:testdataset}
\vspace{-1.5em}
\end{table*}
\noindent {where the importance sampling ratio is computed by }
\begin{equation}
{\rho_{i,t} = \frac{\pi_\theta(o_{i,t} \mid S_0, o_{i,<t})}{\pi_{\theta_{\text{old}}}(o_{i,t} \mid S_0, o_{i,<t})},}
\end{equation}
{and the advantage estimation is updated by }
\begin{equation}
\label{eq:advantage_function}
{A_{i,t}= r_{i,t}-\frac{\sum_{i=1}^{N_\text{rollout}} r_{i,t}}{N_\text{rollout}},}
\end{equation}
where $\mathcal{D}$ is the dataset, $S_0$ represents the initial state for a given rollout, and $O=\{o_i\}_{i=1}^{N_\text{rollout}}$ denotes the set of generated rollouts. $\text{clip}(\cdot)$ denotes the clipping function to balance exploration ($\epsilon_{\text{high}}$) and exploitation ($\epsilon_{\text{low}}$), and $\beta_{\text{KL}}$ denotes the KL regularization hyperparameter.
As shown in Eq. (\ref{eq:grpo_loss}), the loss function consists of two parts: one calculates the policy gradient based on relative advantages, encouraging advantageous tokens and suppressing disadvantageous ones. 
The other, a KL regularization term, constrains the model from diverging excessively from the reference policy, ensuring stable training progress.

More notably, in a key divergence from the standard GRPO algorithm \cite{guo2024deepseekmath}, we omit standard deviation normalization for advantage estimates as shown in Eq.~(\ref{eq:advantage_function}), as it is unstable in our traffic simulation setting. High per-iteration computational costs limit group sizes, and reward masking for motion segments without ground truth yields unreliable batch-level variance estimates. Instead, we normalize advantages using only the mean reward of the motion tokens at each timestep, resulting in more robust learning.
\begin{figure}[!t]

    \includegraphics[width=8.5cm]{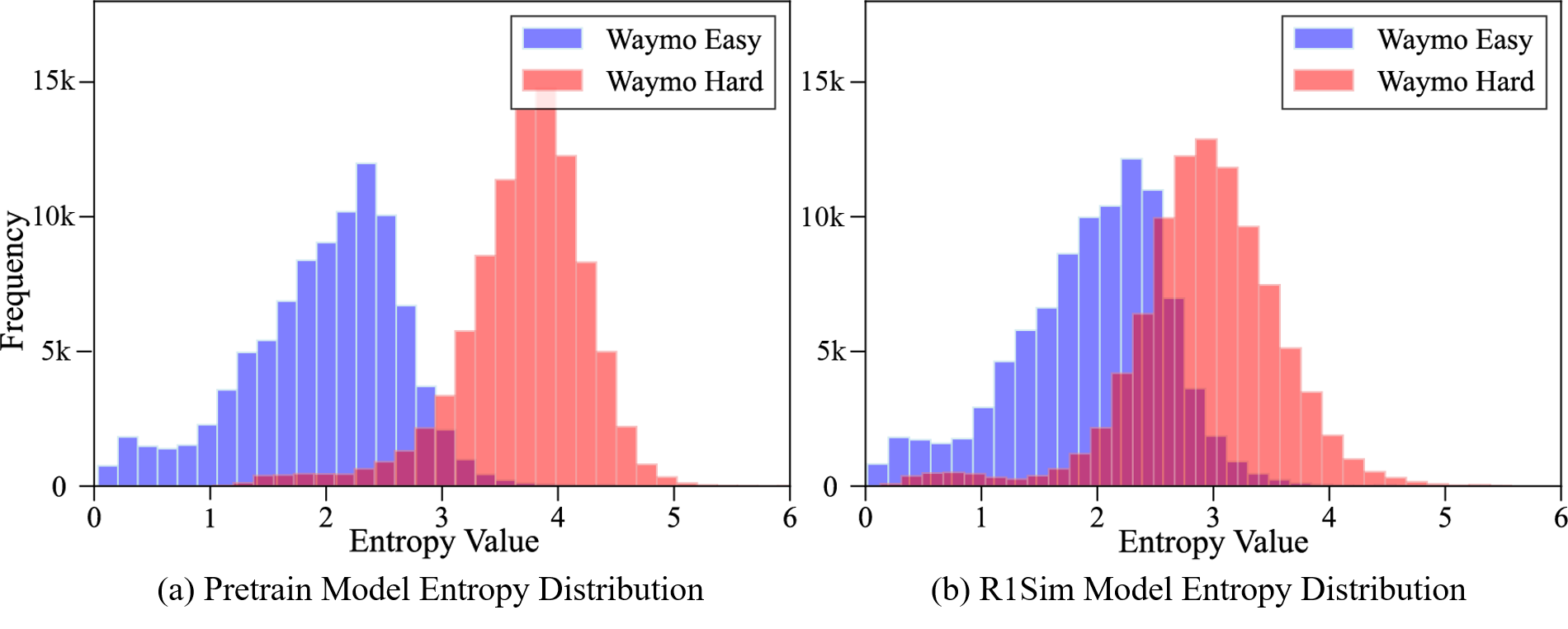}
    \captionsetup{justification=raggedright,singlelinecheck=false} 
    \caption{\textbf{Entropy distribution of generated scenarios.}}
        \label{fig:entropy_distribution_comparison}
\end{figure}
\section{Experiments}
\subsection{Experimental Setup}
\noindent \textbf{Dataset.}
By referring to \cite{montali2023waymo}, we validate the efficacy of our approach by using the Waymo Open Motion Dataset (WOMD) \cite{ettinger2021large}, a large-scale benchmark for traffic simulation containing 486,995/44,097/44,920 training/validation/testing scenarios. Each scenario extends the historic and prediction horizons to 1.1 seconds and 8 seconds, respectively, sampled at 10Hz. 

\noindent \textbf{Implementation Details.}
We consider the open-sourced SOTA algorithms, such as SMART and CATK, as the pretrained model and further finetune the model with our proposed R1Sim for only 1 epoch. To enable training efficiency, the map and agent encoders are frozen except for the final layer. For achieving diverse motion generation, the proposed R1Sim roll outs 32 driving scenarios given the input observations. For both pretraining and fine-tuning, we utilize 4 NVIDIA RTX 4090 GPUs, with a total batch size of 16.

\noindent \textbf{Evaluation Metrics.}
Following \cite{montali2023waymo}, we adopt a set of well-established metrics to assess different aspects of realism.
Specifically, we utilize the approximate negative log likelihood (NLL) to quantify the distributional match between original and simulated scenarios across three critical dimensions: kinematics (e.g., speed and acceleration), interactions (e.g., collision), and map-based (e.g., off-road).
These dimensions are aggregated into the “Realism Meta Metric (RMM)”, a weighted composite score serving as a comprehensive indicator of simulation quality.
\begin{table}[!t]
  \caption{{Samplings and Optimizations Comparison.}}
  \label{tab:WOMD_Ablation_All}
  \centering
  \setlength{\tabcolsep}{1mm}
  \begin{tabular}{l|l|cc|cc}
    \toprule
    \multirow{2}{*}{Category} & \multirow{2}{*}{Method} &
    \multicolumn{2}{c|}{Waymo Easy} &
    \multicolumn{2}{c}{Waymo Hard} \\
    & & RMM & $\Delta$RMM & RMM & $\Delta$RMM \\
    \midrule
    \multirow{4}{*}{Sampling} 
      & Top-K=5      & 0.9039 & -0.0001 & 0.4652 & \textbf{+0.1039} \\
      & Top-K=32     & 0.9018 & \textbf{+0.0020} & 0.5354 & \textbf{+0.0337} \\
      & Top-K=64     & 0.9005 & \textbf{+0.0033} & 0.5591 & \textbf{+0.0100} \\
      & Ours & \textbf{0.9038} & - & \textbf{0.5691} & - \\
    \midrule
    \multirow{3}{*}{Optimization} 
      & SMART (IL)     & 0.8958 & \textbf{+0.0080} & 0.5489 & \textbf{+0.0202} \\
      & CATK (SFT)     & 0.9002 & \textbf{+0.0036} & 0.5595 & \textbf{+0.0096} \\
      & Ours (GRPO)    & \textbf{0.9038} & - & \textbf{0.5691} & - \\
    \bottomrule
  \end{tabular}
\end{table}
\begin{table*}[!t]
  \centering
  \caption{Ablation Study on WOMD Validation Split.}
  \label{tab:ablation}
  \begin{tabular}{l|cc|cccc}
    \toprule
    \multirow{1}{*}{\textbf{Category}} & \multirow{1}{*}{\textbf{Sampling}} & \multirow{1}{*}{\textbf{Optimization}} & RMM (↑) & Kinematic L.(↑) & Interactive L.(↑) & Map-based L.(↑) \\
    \midrule
    Baseline & Top-K=32 & IL & 0.7654($\pm$0.02) & 0.4852($\pm$0.01) & 0.8056($\pm$0.01) & 0.8737($\pm$0.02) \\
    \midrule
    \multirow{3}{*}{\textbf{+ Sampling}} 
     & Top-K=5 & IL & 0.7545($\pm$0.01) & 0.4554($\pm$0.00) & 0.8013($\pm$0.01) & 0.8652($\pm$0.01) \\
     & Top-K=64 & IL & 0.7629($\pm$0.03) & 0.4855($\pm$0.01) & 0.8005($\pm$0.00) & 0.8730($\pm$0.03) \\
     & Entropy-guided & IL & 0.7670($\pm$0.01) & 0.4870($\pm$0.01) & 0.8076($\pm$0.01) & 0.8747($\pm$0.02) \\
    \midrule
    \multirow{3}{*}{\textbf{+ Optimization}} 
     & Top-K=32 & SFT & 0.7669($\pm$0.01) & 0.4862($\pm$0.01) & 0.8079($\pm$0.01) & 0.8745($\pm$0.02) \\
     & {Top-K=32} & {GRPO-standard} & {0.7619($\pm$0.02)} & {0.4729($\pm$0.01)} & {0.8035($\pm$0.01)} & {0.8736($\pm$0.03)} \\
     & {Top-K=32} & {GRPO-refined} & {0.7675($\pm$0.03)} &{ \textbf{0.4883}($\pm$0.02)} & {0.8079($\pm$0.00)} &{0.8751($\pm$0.02)} \\
    \midrule
    \textbf{Both (Ours)} & Entropy-guided & GRPO-refined & \textbf{0.7683}($\pm$0.02) & 0.4878($\pm$0.01) & \textbf{0.8090}($\pm$0.00) & \textbf{0.8763}($\pm$0.02) \\
    \bottomrule
  \end{tabular}
  \vspace{-1.5em}
\end{table*}
\subsection{Results on Benchmarks}
As shown in TABLE I, we conduct comprehensive comparisons in the WOMD test set and achieve competitive results against baselines. Compared with early tokenized representatives, our proposed R1Sim demonstrates superior motion generation performance across almost all evaluation aspects, exhibiting its ability to achieve optimal decision-making that closely aligns with human preference.

To further evaluate the generality of our model, we perform the proposed GRPO built upon the pretrained model from the SOTA method SMART and the latest advanced SFT method CATK. Comparison results show that our proposed R1Sim enhances the motion generation performance of SMART, particularly in critical metrics, including interactive metrics, kinematic metrics, and map-based metrics. In contrast with SMART-tiny, our proposed method enables safer multi-agent motion generation with an increased 0.0117 of the collision likelihood metric. Therefore, promising performance can be attributed to the expanded explorations of potentially optimal motion tokens using GRPO.

\subsection{Comparison in Different Uncertainty Scenes}

This section investigates whether R1Sim adaptively benefits scenarios with different uncertainty levels. We split the validation dataset into Waymo Easy and Waymo Hard subsets based on the pretrained model’s performance, by selecting the top and bottom 2\% of scenarios ranked by RMM, respectively. This split reflects cases where the pretrained policy performs particularly well or poorly. We then analyze the token entropy distributions within each subset. As shown in Fig. 4, scenarios with lower RMM scores consistently exhibit higher token entropy, revealing a strong correlation between degraded generation quality and elevated uncertainty in token selection. This observation indicates that token entropy serves as an effective proxy for scene difficulty and modeling uncertainty. In contrast, R1Sim substantially reduces token entropy in both subsets and, more importantly, narrows the entropy gap between easy and hard scenarios. 
This implies that R1Sim goes beyond uniform performance gains to actively address uncertainty in difficult scenes, primarily by directing more exploration capacity toward high-entropy motion patterns.

\noindent \textbf{Entropy-conditioned Sampling Comparison.} We investigate the impact of different sampling strategies on model performance across scene types with varying levels of token entropy. As shown in TABLE \ref{tab:WOMD_Ablation_All}, we observe that in low-entropy scenes, increasing the sampling size K degrades motion generation performance, since agent behavior is largely deterministic and excessive sampling introduces unnecessary noise. In contrast, in high-entropy and more challenging scenes, larger K values significantly improve performance by enabling exploration over a more diverse set of candidate tokens. This increased diversity facilitates the discovery of “hidden gem” motion trajectories that can be recovered from suboptimal initial states and is essential for achieving high realism in complex scenarios. These findings provide strong empirical support for the theoretical motivation of our proposed R1Sim framework. By accounting for uncertainty in agent motion through entropy-guided adaptive sampling, R1Sim dynamically adjusts the sampling range over time, achieving consistent performance improvements across scenes with varying complexity.

\noindent \textbf{Entropy-conditioned Optimization Comparison.} Uniformly built upon the entropy-guided adaptive sampling, we further explore the effectiveness of our proposed GRPO with different scenario complexities. As shown in TABLE \ref{tab:WOMD_Ablation_All}, our method consistently enhances overall performance against the IL baseline SMART and SFT baseline CATK. Specifically in the high-entropy scenes, our method achieves improvements of 0.0202 and 0.0096 RMM. These margins are particularly significant given the overall lower baseline performance in such complex scenarios, highlighting the advantage of GRPO in uncertainty-aware decision making. Instead of purely behavior mimicking, these promising results can be attributed to our method with the safety-aware reward, effectively discovering “hidden gem” token sequences for generating human-preferred  behaviors.

\begin{figure}[!t]

    \includegraphics[width=8.5cm]{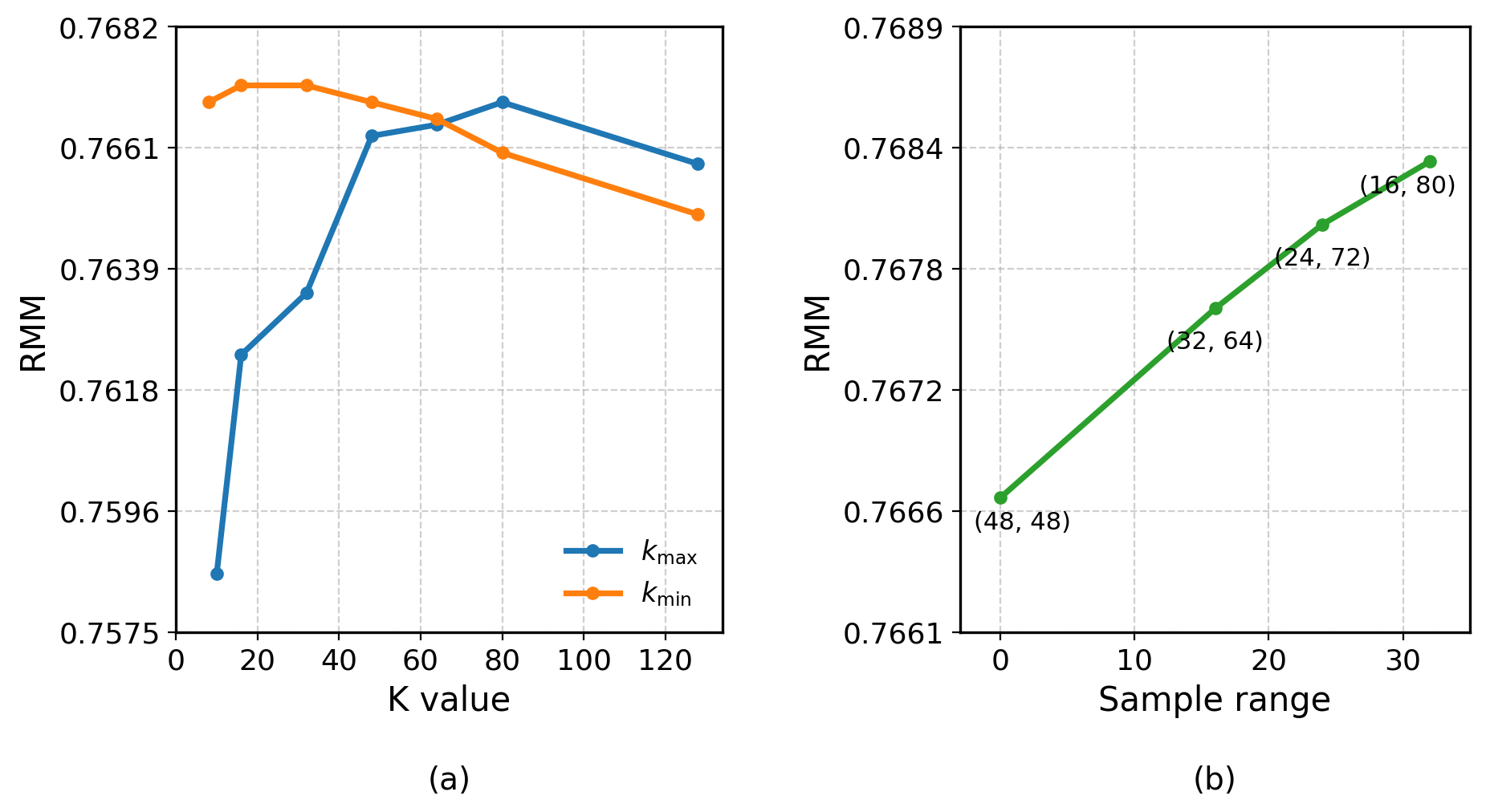}
    \captionsetup{justification=raggedright,singlelinecheck=false} 
    \caption{\textbf{{Impact of the minimum bound $k_{min}$, maximum bound $k_{max}$ and sample ranges on RMM.}}}
        \label{fig:Sensitivity}
\vspace{-1em}
\end{figure}
\begin{figure*}[!t]
	\centering
	\begin{center}
		\scriptsize
		\includegraphics*[width=\textwidth]{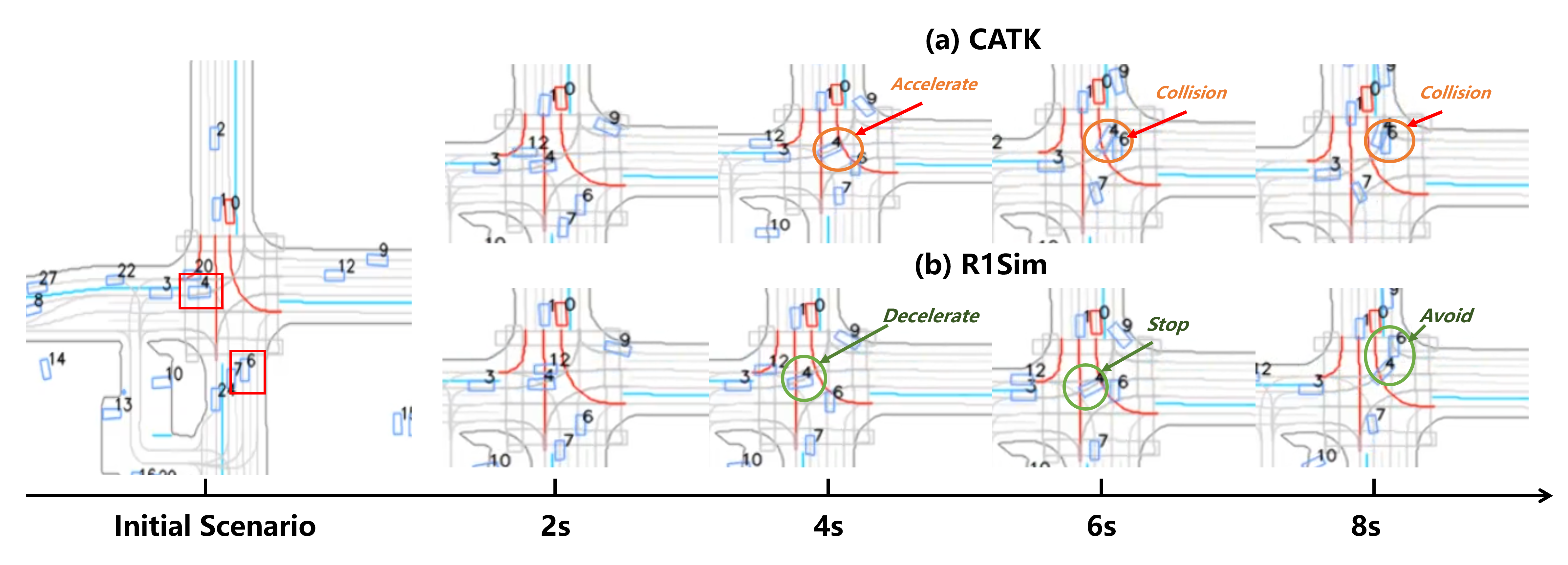}
		\caption{\textbf{{Qualitative comparison of closed-loop rollouts.}} {The left panel displays the initial scenario. The right panels illustrate the temporal evolution generated by (a) CATK\cite{zhang2025closed} and (b) our R1Sim. Red boxes highlight the interested agents (ID 4 and 6), while circles (orange/green) mark critical interaction behaviors.}}
		\label{fig:qualitative}
	\end{center}
\vspace{-1em}
\end{figure*}
\subsection{Ablation Study}

In this section, we conduct ablation studies for our proposed R1Sim on the WOMD validation split, and demonstrate how our architecture choices affect the model performance. {During ablations, we take SMART\cite{wu2024smart} as the baseline model and progressively incorporate additional components into the baseline. For each experimental setting, we conduct five independent runs and report the mean performance to ensure the stability and reliability of the results as shown in TABLE~\ref{tab:ablation}}.

\noindent \textbf{Impact of Sampling Design.} Compared to fixed sampling strategies such as Top-K, our proposed entropy-guided adaptive sampling strategy effectively selects the range of plausible motion tokens according to scene-specific uncertainty, leading to a substantial improvement in motion realism. By tailoring the sampling space to the underlying entropy of the scene, the model avoids unnecessary noise in deterministic scenarios while enabling sufficient exploration in more complex ones.

\noindent \textbf{Impact of Optimization Design.} Our proposed GRPO enhances the value estimation of motion tokens, aligning the token selection process with reward functions that reflect human preferences. This alignment leads to notable improvements in RMM and guides the model to generate motions that better conform to human behavioral expectations. We further compare the standard GRPO formulation (GRPO-standard) with our refined variant that removes the standard deviation term (GRPO-refined). The results show that eliminating the standard deviation stabilizes training and yields consistent performance gains. Finally, the synergistic outcome implies that our adaptive sampling strategy, by concentrating exploration in high-entropy while high-uncertainty regions, effectively enables the GRPO.

\subsection{Sensitivity Analysis} 
\noindent \textbf{Sensitivity to Entropy-guided Sampling Regions.} We analyze the sensitivity of the proposed entropy-guided adaptive sampling strategy with respect to its key hyperparameters, including the minimum and maximum bounds $k_{\min}$ and $k_{\max}$ , as well as the sampling ranges. As shown in Fig.~\ref{fig:Sensitivity} (a), the increase of $k_{\max}$ improves motion simulation performance by enabling sufficient exploration in high-entropy motion patterns, while overly large values introduce unnecessary noise and marginally degrade performance. In contrast, larger values of $k_{\min}$ consistently harms performance by restricting adaptivity and forcing over-exploration in low-entropy scenarios. As shown in Fig.~\ref{fig:Sensitivity} (b), expanding the sampling ranges further improves realism by allowing a broader set of plausible behaviors to be explored. Based on these observations, we adopt ${(16,80)}$, as a balanced default setting that reconciles accuracy and exploratory flexibility.
\begin{table}[!t]
  \centering
        \caption{{Impact of Reward Designs.}}
  \small
    \begin{tabular}{r|cccc}
    \toprule
    \multicolumn{1}{c|}{ID} & 
    \begin{tabular}[c]{@{}c@{}}RMM~(↑)\end{tabular} & \begin{tabular}[c]{@{}c@{}}Kinematic\\L.~(↑)\end{tabular} &  \begin{tabular}[c]{@{}c@{}}Interactive\\L.~(↑)\end{tabular} & \begin{tabular}[c]{@{}c@{}}Map-based\\L.~(↑)\end{tabular} \\
    \midrule
    OR  & 0.7635 & 0.4798 & 0.8070 & 0.8698 \\
    APR  & 0.7644 & 0.4858 & 0.8052 & 0.8714 \\
    AHR  & 0.7656 & 0.4855 & 0.8056 & 0.8741 \\
    SHR  & 0.7668   & 0.4871   & 0.8077   &  0.8741 \\
    \midrule
    SPR  &  \textbf{0.7683}  &  \textbf{0.4878}  &\textbf{0.8090}   &  \textbf{0.8763}\\
    \bottomrule
    \end{tabular}%
  \label{tab:reward}%
\end{table}%

\noindent\textbf{Sensitivity to Reward Designs.} We conduct a comparative analysis of several reward formulations, including the outcome reward, the process reward, and the hybrid reward, to evaluate the effectiveness of the proposed safety-aware process reward. Specifically, the outcome reward (OR) assesses trajectory quality only at the final step by aggregating kinematic and collision terms, while the process reward (PR) provides step-wise supervision during rollouts. The PR includes an additive variant (APR), which sums safety reward and realism reward, and a safety-aware variant (SPR), which uses the safety term as a multiplicative weighting on the distance penalty. We further consider hybrid rewards (HR) that combine outcome and process supervision, resulting in additive (AHR) and safety-aware (SHR) hybrids. As shown in TABLE~\ref{tab:reward}, SPR consistently achieves the best performance across all metrics and yields the highest RMM. These results indicate that process rewards offer more effective dense supervision for long-horizon generation than sparse outcome rewards. More notably, incorporating safety as a multiplicative factor provides stronger and more stable gradient guidance than additive designs APR. {In addition, SPR outperforms SHR across all evaluation metrics, suggesting that the safety-aware process reward alone is sufficient without relying on redundant outcome-level signals. Our proposed safety-aware process reward design efficiently balances the relationship between safety and fidelity, thereby reducing the burden of parameter tuning.}

\section{Qualitative results}

{We conducte a qualitative comparison between R1Sim and the SOTA baseline CATK on the WOMD validation set. As illustrated in Fig. \ref{fig:qualitative}, we visualize the multi-step rollout trajectories generated by both models.}

In the CATK rollout, the model exhibits an overly aggressive driving policy. As highlighted by the orange ellipse, the left-turning vehicle accelerates in an attempt to cut through the traffic, failing to anticipate the oncoming vehicle and resulting in a collision. In contrast, R1Sim demonstrates rational and safe behavior. The green ellipse shows the agent proactively decelerating upon observing the straight-going vehicle, effectively yielding the right-of-way before completing the merge. These results show that R1Sim goes beyond simple imitation, which deeply comprehends the underlying logic of driving interactions in complex scenes and enables reasonable yielding maneuvers that guarantee safety.
\section{Conclusion}
In this paper, we introduce R1Sim, a novel tokenized traffic simulation framework that pioneers an R1-style post-training paradigm to effectively balance exploration and exploitation. Recognizing that token entropy reflects underlying motion uncertainty, we propose an entropy-guided adaptive sampling strategy to uncover previously overlooked, high-potential driving behaviors. To ensure realism and safety, these behaviors are further refined using Group Relative Policy Optimization (GRPO) guided by safety-aware rewards. Evaluations on the WOMD Sim Agent benchmark demonstrate that R1Sim achieves competitive performance, successfully delivering diverse and human-preferred motion generation.

\ifCLASSOPTIONcaptionsoff
  \newpage
\fi

\bibliographystyle{IEEEtran}
\bibliography{IEEEabrv,Bibliography}

@article{achiam2023gpt,
  title={Gpt-4 technical report},
  author={Achiam, Josh and Adler, Steven and Agarwal, Sandhini and et al.},
  year={2024},
  journal={arXiv preprint arXiv:2303.08774},
}

@article{guo2024deepseekmath,
  title={Deepseekmath: Pushing the limits of mathematical reasoning in open language models},
  author={Shao, Zhihong and Wang, Peiyi and Zhu, Qihao and et al.},
  year={2024},
  journal={arXiv preprint arXiv:2402.03300},
}

@inproceedings{zhang2025closed,
  title = {Closed-Loop Supervised Fine-Tuning of Tokenized Traffic Models},
  author = {Zhang, Zhejun and Karkus, Peter and Igl, Maximilian and et al.},
  booktitle = {Proceedings of the IEEE Conference on Computer Vision and Pattern Recognition (CVPR)},
  pages={5422--5432},
  year = {2025},
}

@inproceedings{zhou2024behaviorgpt,
title={Behavior{GPT}: Smart Agent Simulation for Autonomous Driving with Next-Patch Prediction},
author={Zikang, Zhou and Haibo, HU and Xinhong, Chen and et al.},
booktitle={Proceedings of the Annual Conference on Neural Information Processing Systems (NeurIPS)},
volume={37},
pages={79597--79617},
year={2024}
}

@inproceedings{wu2024smart,
  title={SMART: Scalable Multi-agent Real-time Simulation via Next-token Prediction},
  author={Wu, Wei and Feng, Xiaoxin and Gao, Ziyan and Kan, Yuheng},
  booktitle={Proceedings of the Annual Conference on Neural Information Processing Systems (NeurIPS)},
  volume={37},
  pages={114048--114071},
  year={2024}
}

@article{philion2024trajeglishtrafficmodelingnexttoken,
      title={Trajeglish: Traffic Modeling as Next-Token Prediction}, 
      author={Jonah Philion and Xue Bin Peng and Sanja Fidler},
      year={2024},
      journal={arXiv preprint arXiv:2312.04535},
}

@inproceedings{seff2023motionlm,
  title={Motionlm: Multi-agent motion forecasting as language modeling},
  author={Seff, Ari and Cera, Brian and et al.},
  booktitle={Proceedings of the IEEE/CVF International Conference on Computer Vision (CVPR)},
  pages={8579--8590},
  year={2023}
}

@inproceedings{hu2024solving,
  title={Solving motion planning tasks with a scalable generative model},
  author={Hu, Yihan and Chai, Siqi and Yang, Zhening and et al.},
  booktitle={Proceedings of the European Conference on Computer Vision (ECCV)},
  pages={386--404},
  year={2024},
}

@article{zhao2024kigras,
  title={Kigras: Kinematic-driven generative model for realistic agent simulation},
  author={Zhao, Jianbo and Zhuang, Jiaheng and Zhou, Qibin and et al.},
  journal={IEEE Robotics and Automation Letters},
  year={2025},
  volume={10},
  number={2},
  pages={1082-1089},
}

@article{lin2025revisit,
  author={L. Lin and X. Lin and K. Xu and others},
  title={Revisit Mixture Models for Multi-Agent Simulation: Experimental Study within a Unified Framework},
  journal={arXiv preprint arXiv:2501.17015},
  year={2025},
}

@inproceedings{lu2023imitation,
  title={Imitation is not enough: Robustifying imitation with reinforcement learning for challenging driving scenarios},
  author={Lu, Yiren and Fu, Justin and Tucker, George and et al.},
  booktitle={Proceedings of the IEEE/RSJ International Conference on Intelligent Robots and Systems (IROS)},
  pages={7553--7560},
  year={2023},
}

@article{li2025finetuning,
  title={Finetuning generative trajectory model with reinforcement learning from human feedback},
  author={Li, Derun and Ren, Jianwei and Wang, Yue and et al.},
  year={2025},
  journal={arXiv preprint arXiv:2503.10434},
}

@inproceedings{zhang2025carplanner,
  title={Carplanner: Consistent auto-regressive trajectory planning for large-scale reinforcement learning in autonomous driving},
  author={Zhang, Dongkun and Liang, Jiaming and Guo, Ke and et al.},
  booktitle={Proceedings of the Computer Vision and Pattern Recognition Conference (CVPR)},
  pages={17239--17248},
  year={2025}
}

@article{huang2024gen,
  title={Gen-Drive: Enhancing Diffusion Generative Driving Policies with Reward Modeling and Reinforcement Learning Fine-tuning},
  author={Huang, Zhiyu and Weng, Xinshuo and Igl, Maximilian and et al.},
  year={2024},
  journal={arXiv preprint arXiv:2410.05582},
}

@inproceedings{ettinger2021large,
  title={Large scale interactive motion forecasting for autonomous driving: The waymo open motion dataset},
  author={Ettinger, Scott and Cheng, Shuyang and Caine, Benjamin and et al.},
  booktitle={Proceedings of the IEEE/CVF International Conference on Computer Vision (CVPR)},
  pages={9710--9719},
  year={2021}
}

@article{wang20258020rulehighentropyminority,
      title={Beyond the 80/20 Rule: High-Entropy Minority Tokens Drive Effective Reinforcement Learning for LLM Reasoning}, 
      author={Shenzhi Wang and Le Yu and Chang Gao and et al.},
      year={2025},
      journal={arXiv preprint arXiv:2506.01939},
}

@article{cui2025entropymechanismreinforcementlearning,
      title={The Entropy Mechanism of Reinforcement Learning for Reasoning Language Models}, 
      author={Ganqu Cui and Yuchen Zhang and Jiacheng Chen and et al.},
      year={2025},
      journal={arXiv preprint arXiv:2505.22617},
}

@article{schulman2017proximal,
  title={Proximal policy optimization algorithms},
  author={Schulman, John and Wolski, Filip and Dhariwal, Prafulla and et al.},
  year={2017},
  journal={arXiv preprint arXiv:1707.06347},
}

@inproceedings{de2019causal,
  title={Causal confusion in imitation learning},
  author={De Haan, Pim and Jayaraman, Dinesh and Levine, Sergey},
  booktitle={Proceedings of the Annual Conference on Neural Information Processing Systems (NeurIPS)},
  volume={32},
  year={2019}
}

@inproceedings{montali2023waymo,
  title={The waymo open sim agents challenge},
  author={Montali, Nico and Lambert, John and et al.},
  booktitle={Proceedings of the Annual Conference on Neural Information Processing Systems (NeurIPS)},
  volume={36},
  pages={59151--59171},
  year={2023}
}

@book{gray2011entropy,
  title={Entropy and information theory},
  author={Gray, Robert M},
  year={2011},
  publisher={Springer Science \& Business Media}
}

@article{fan2018hierarchical,
  title={Hierarchical neural story generation},
  author={Fan, Angela and Lewis, Mike and Dauphin, Yann},
  year={2018},
  journal={arXiv preprint arXiv:1805.04833},
}

@article{meister2025locallytypicalsampling,
  title={Locally typical sampling},
  author={Meister, Clara and Pimentel, Tiago and Wiher, Gian and Cotterell, Ryan},
  journal={Transactions of the Association for Computational Linguistics},
  volume={11},
  pages={102--121},
  year={2023}
}

@article{zhao2025dropedirectionalrotaryposition,
      title={DRoPE: Directional Rotary Position Embedding for Efficient Agent Interaction Modeling}, 
      author={Jianbo Zhao and Taiyu Ban and Zhihao Liu and et al.},
      year={2025},
      journal={arXiv preprint arXiv:2503.15029},
}

@article{guo2025deepseekr1,
  title={Deepseek-r1 incentivizes reasoning in llms through reinforcement learning},
  author={Guo, Daya and Yang, Dejian and Zhang, Haowei and et al.},
  journal={Nature},
  volume={645},
  number={8081},
  pages={633--638},
  year={2025},
  publisher={Nature Publishing Group UK London}
}

@article{feng2021intelligent,
  title={Intelligent driving intelligence test for autonomous vehicles with naturalistic and adversarial environment},
  author={Feng, Shuo and Yan, Xintao and Sun, Haowei and et al.},
  journal={Nature communications},
  volume={12},
  number={1},
  pages={748},
  year={2021},
  publisher={Nature Publishing Group UK London}
}

@article{luo2022gamma,
  title={GAMMA: A general agent motion model for autonomous driving},
  author={Luo, Yuanfu and Cai, Panpan and Lee, Yiyuan and Hsu, David},
  journal={IEEE Robotics and Automation Letters},
  volume={7},
  number={2},
  pages={3499--3506},
  year={2022},
  publisher={IEEE}
}

@inproceedings{yang2025drivearena,
  title={Drivearena: A closed-loop generative simulation platform for autonomous driving},
  author={Yang, Xuemeng and Wen, Licheng and Wei, Tiantian and et al.},
  booktitle={Proceedings of the IEEE/CVF International Conference on Computer Vision},
  pages={26933--26943},
  year={2025}
}

@article{jiang2025alphadrive,
  title={Alphadrive: Unleashing the power of vlms in autonomous driving via reinforcement learning and reasoning},
  author={Jiang, Bo and Chen, Shaoyu and Zhang, Qian and et al.},
  journal={arXiv preprint arXiv:2503.07608},
  year={2025}
}

@article{dou2025plan,
  title={Plan Then Action: High-Level Planning Guidance Reinforcement Learning for LLM Reasoning},
  author={Dou, Zhihao and Zhao, Qinjian and Wan, Zhongwei and et al.},
  journal={arXiv preprint arXiv:2510.01833},
  year={2025}
}

@inproceedings{parmar2025plan,
  title={Plan-tuning: Post-training language models to learn step-by-step planning for complex problem solving},
  author={Parmar, Mihir and Goyal, Palash and Liu, Xin and et al.},
  booktitle={Proceedings of the 2025 Conference on Empirical Methods in Natural Language Processing},
  pages={21430--21444},
  year={2025}
}

@article{tang2025plan,
  title={Plan-R1: Safe and Feasible Trajectory Planning as Language Modeling},
  author={Tang, Xiaolong and Kan, Meina and Shan, Shiguang and Chen, Xilin},
  journal={arXiv preprint arXiv:2505.17659},
  year={2025}
}

\end{document}